\makeatletter
\declare@file@substitution{acmart.cls}{styles/vendor/acmart/acmart.cls}
\makeatother
\documentclass[sigconf,nonacm]{acmart}

\usepackage{tabularx}
\usepackage{multirow}
\usepackage{enumitem}
\usepackage{xspace}
\usepackage{subcaption}
\usepackage{balance}
\usepackage{placeins}

\setcopyright{none}
\hypersetup{hidelinks,keeppdfinfo}
\newcommand{\ATPKeywords}{Agentic Web, Agent-to-Agent Communication, AI Agents, Agent Transfer Protocol, Server-Mediated Communication}
\newcommand{\ATPAssetRoot}{assets/figures}
\newcommand{\ATPBibliographyStyle}{styles/vendor/acmart/ACM-Reference-Format}

\newcommand{\ATPEmailYuqiQiu}{qiuyuqi@nankai.edu.cn}
\newcommand{\ATPEmailZuyaoXu}{xuzuyao@nankai.edu.cn}
\newcommand{\ATPEmailLuSun}{2112060@nankai.edu.cn}
\newcommand{\ATPEmailXiangLi}{lixiang@nankai.edu.cn}
\newcommand{\ATPEmailFashengMiao}{mfs24@mails.tsinghua.edu.cn}
\newcommand{\ATPEmailFubinWu}{yuyinqiyue@gmail.com}
\newcommand{\ATPEmailHaixinDuan}{duanhx@tsinghua.edu.cn}
\newcommand{\ATPEmailZheliLiu}{liuzheli@nankai.edu.cn}
\newcommand{\ATPNankaiInstitution}{Nankai University}
\newcommand{\ATPTsinghuaInstitution}{Tsinghua University}
\newcommand{\ATPQuanchengInstitution}{Quancheng Laboratory}
\newcommand{\ATPNankaiCity}{Tianjin}
\newcommand{\ATPTsinghuaCity}{Beijing}
\newcommand{\ATPQuanchengCity}{Jinan}
\newcommand{\ATPCountry}{China}

\newcommand{\ATPAuthorFont}{\small\normalfont}
\newcommand{\ATPTableFont}{\small}
\newcommand{\ATPParagraphSkip}{\smallskipamount}
\newcommand{\ATPParagraphLabelGap}{0.5em}
\newcommand{\ATPCaptionSkip}{4pt}
\newcommand{\ATPSubcaptionSkip}{2pt}
\newcommand{\ATPColumnPanelWidth}{0.485\linewidth}
\newcommand{\ATPWidePanelWidth}{0.32\textwidth}
\newcommand{\ATPMessagePanelWidth}{0.26\textwidth}
\newcommand{\ATPRequestPanelWidth}{0.38\textwidth}
\newcommand{\ATPSubscriptionPanelWidth}{0.33\textwidth}
\newcommand{\ATPFlowHeight}{35mm}
\newcommand{\ATPCompactTableColumnSep}{3pt}
\newcommand{\ATPWideTableColumnSep}{2pt}
\newcommand{\ATPWideTableArrayStretch}{1.12}
\newcommand{\ATPCheckLabelWidth}{0.30\columnwidth}
\newcommand{\ATPResponseLabelWidth}{0.36\columnwidth}
\newcommand{\ATPResponseActionWidth}{0.27\columnwidth}
\newcommand{\ATPAttackCaseWidth}{0.14\textwidth}
\newcommand{\ATPAttackConstructionWidth}{0.3\textwidth}
\newcommand{\ATPAttackCheckWidth}{0.15\textwidth}
\newcommand{\ATPAttackInboxWidth}{0.10\textwidth}

\newcommand{\sys}{ATP\xspace}
\newcommand{\para}[1]{%
  \par\addvspace{\ATPParagraphSkip}\noindent\textbf{#1}\hspace{\ATPParagraphLabelGap}}

\makeatletter
\renewcommand{\@affiliationfont}{\ATPAuthorFont}
\makeatother

\newcommand{\ATPCompactFigureCaptions}{%
  \captionsetup{skip=\ATPCaptionSkip}%
  \captionsetup[subfigure]{skip=\ATPSubcaptionSkip}}

\newcommand{\ATPCompactTable}{%
  \ATPTableFont\setlength{\tabcolsep}{\ATPCompactTableColumnSep}}
\newcommand{\ATPWideTable}{%
  \ATPTableFont\setlength{\tabcolsep}{\ATPWideTableColumnSep}%
  \renewcommand{\arraystretch}{\ATPWideTableArrayStretch}}

\makeatletter
\patchcmd{\@BAlancecol}
  {\setbox\@leftcolumn=\vbox to \@tempdima}
  {\advance\@tempdima by \dp1 \setbox\@leftcolumn=\vbox to \@tempdima}
  {}{\PackageWarning{atp-preprint}{Could not apply balance depth correction}}
\makeatother

\title{Agent Transfer Protocol: Server-Mediated Agent-to-Agent Communication Architecture for the Agentic Web}

\author{Yuqi Qiu}
\expandafter\email\expandafter{\ATPEmailYuqiQiu}
\affiliation{%
  \institution{\ATPNankaiInstitution}
  \city{\ATPNankaiCity}
  \country{\ATPCountry}}

\author{Zuyao Xu}
\expandafter\email\expandafter{\ATPEmailZuyaoXu}
\affiliation{%
  \institution{\ATPNankaiInstitution}
  \city{\ATPNankaiCity}
  \country{\ATPCountry}}

\author{Lu Sun}
\expandafter\email\expandafter{\ATPEmailLuSun}
\affiliation{%
  \institution{\ATPNankaiInstitution}
  \city{\ATPNankaiCity}
  \country{\ATPCountry}}

\author{Xiang Li}
\authornote{Corresponding author.}
\expandafter\email\expandafter{\ATPEmailXiangLi}
\affiliation{%
  \institution{\ATPNankaiInstitution}
  \city{\ATPNankaiCity}
  \country{\ATPCountry}}

\author{Fasheng Miao}
\expandafter\email\expandafter{\ATPEmailFashengMiao}
\affiliation{%
  \institution{\ATPTsinghuaInstitution}
  \city{\ATPTsinghuaCity}
  \country{\ATPCountry}}

\author{Fubin Wu}
\expandafter\email\expandafter{\ATPEmailFubinWu}
\affiliation{%
  \institution{\ATPNankaiInstitution}
  \city{\ATPNankaiCity}
  \country{\ATPCountry}}

\author{Haixin Duan}
\expandafter\email\expandafter{\ATPEmailHaixinDuan}
\affiliation{%
  \institution{\ATPTsinghuaInstitution}
  \city{\ATPTsinghuaCity}
  \country{\ATPCountry}}
\affiliation{%
  \institution{\ATPQuanchengInstitution}
  \city{\ATPQuanchengCity}
  \country{\ATPCountry}}

\author{Zheli Liu}
\expandafter\email\expandafter{\ATPEmailZheliLiu}
\affiliation{%
  \institution{\ATPNankaiInstitution}
  \city{\ATPNankaiCity}
  \country{\ATPCountry}}

\renewcommand{\shortauthors}{Qiu et al.}

\begin{document}

\begin{abstract}
The Agentic Web needs to support continuous collaboration among agents from different organizations.
These collaborative relationships may span multiple tasks, network connections and execution instances.
Such interactions require stable communication entries, message reception when instances are offline, and source verification after delayed reading.
How to organize these responsibilities so that cross domain communication does not depend on specific execution instances being continuously online is a key issue for building long term agent interaction infrastructure.

We propose ATP (Agent Transfer Protocol), a cross organization agent communication protocol adopting a domain mediated architecture.
In this architecture, administrative domains maintain stable addresses and receiving entries for logical agents.
The domain servers of both parties continuously undertake cross domain transmission.
The receiving domain verifies and saves messages for execution instances to read later.
The sending domain signs the complete message envelope.
This allows the receiving instance to independently reverify the domain source and content integrity after the original connection ends, provided the original verification public key is still trusted and obtainable.
We implemented an ATP prototype and evaluated it through a controlled cross organization LLM agent workflow, mechanism checks and latency measurements.
The workflow completed subsequent business processing after instances went offline or were replaced.
The independent verifier successfully verified the domain source and content integrity of saved messages.
Latency measurements further quantified the communication and processing overhead introduced by the continuous mediation of domain servers.

\end{abstract}

\ccsdesc[500]{Networks~Application layer protocols}
\ccsdesc[300]{Security and privacy~Network security}

\keywords{\ATPKeywords}

\maketitle
\pagestyle{plain}

\section{Introduction}
\label{sec:intro}

Large language model (LLM)-based agents are software entities that autonomously plan actions, invoke tools, and interact with other agents to achieve user goals \cite{shavit2023practices,sander2026taxonomy}.
In the vision of the Agentic Web, agents developed and operated by diverse entities will represent users, communicating and collaborating autonomously across service, platform, and organizational boundaries \cite{yang2025agenticweb,petrova2025webofagents}.
For instance, a travel agent can collaborate with agents operated by airlines and hotels to complete an itinerary.
To realize this cross-domain collaboration, systems require a shared communication infrastructure that is independent of internal agent implementations.
Such an infrastructure needs to provide common addressing methods and messaging protocols, while supporting participating parties in verifying peer identities and delegated permissions \cite{chan2025infrastructure,south2025position}.
Cross-organizational interactions can be long-lived and span multiple intermediaries and trust boundaries.
This makes communication continuity and reliability a fundamental requirement for cross-platform and cross-vendor interoperability.
When participants change devices or network attachments, interactions are interrupted, or network paths and intermediate nodes fail, the communication infrastructure needs to maintain the association among related interactions, support failure recovery, and handle temporary peer unreachability.
These system requirements are explicitly included in the target scope of the dialog management protocol and reference architecture for the recently chartered IETF AgentProto working group \cite{agentproto-charter}.

Long-lived interactions often persist through the restart, migration, or replacement of participant execution instances.
The externally identifiable communicating entity, termed a logical agent, can be represented by different execution instances either sequentially or concurrently \cite{bu2026principalbinding}.
While its identity and communication relationships span multiple tasks, the underlying processes, devices, and network connections may change.
Each administrative domain independently maintains local naming, identities, and access policies for its logical agents, and manages their respective execution instances.
External participants only need to interact with the logical entity, without tracking instance-level changes within the target domain.
This dynamic requires domains to coordinate inter-domain interactions through shared rules while independently managing their local deployments \cite{chan2025infrastructure}.
Specifically, these rules need to define the addressing methods and message reception responsibilities for logical entities, and formalize the mechanisms for receivers to verify provenance endorsements and their asserting domains after messages have been stored, forwarded, or delivered with a delay.

In schemes using service endpoints as interaction boundaries, the above requirements still need to be achieved through service deployment and additional interdomain agreements.
Based on A2A, combining continuously running receiving services, persistent queues and message signatures can support communication of logical agents during instance changes \cite{a2a-spec,rfc9421}.
Even if identity and discovery are provided by shared services \cite{syros2026saga,anp-whitepaper,macp-draft} and entry authentication is executed by managed gateways \cite{aws-agentcore-runtime}, deployers still need to clarify how logical addresses associate with receiving entries, when reception confirmation represents the message is saved and custody responsibility transfer is completed, and how verification keys bind with the sending domain in the message.
These agreements need to be jointly followed by both cross domain parties.
They cannot be determined solely by the internal implementation of one party.
The administrative domain provides a natural boundary to organize these agreements.
It maintains local naming, identity credentials and access policies.
It can put logical addresses, message reception and domain level source endorsement under the same management authority.
Meanwhile it keeps execution instance changes processed within the domain.

We propose ATP (Agent Transfer Protocol), a cross organization agent communication protocol adopting a domain mediated architecture.
Inspired by the email architecture, ATP uses each administrative domain to maintain addresses like \texttt{local@domain} for logical agents (where \texttt{local} is the intra domain identifier of the agent, and \texttt{domain} is the domain name of the administrative domain it belongs to).
After authenticating the local sender, the sending domain signs the complete message envelope and locates the receiving domain entry through DNS.
After verifying and saving the message, the receiving domain confirms reception and assumes custody responsibility.
Instances authenticated by this domain can then read the message within the retention period.
The same logical agent can continue to use the original address after instance replacement.
The instance also does not need to expose public network inbound endpoints.
The signature is saved with the message.
When the original public key is still trusted and obtainable, the receiving instance can reverify the domain source and content integrity after the connection ends.
ATP provides interdomain message transmission and source verification.
Task status and dialog semantics can be organized by upper layer protocols like A2A.
Whether specific business operations are authorized is judged by the receiving application according to local policies and related delegation credentials.

Our contributions are as follows:

\textbf{Problem definition of continuous cross organization communication.} We distinguish logical agents, execution instances, tasks and network connections.
We characterize the requirements when communication relationships cross task ends, connection interruptions and instance changes.
We clarify the infrastructure responsibilities involved in stable addressing, continuous reception and source verification after saving.

\textbf{Domain mediated architecture and ATP implementation.} We design a communication architecture where the domain servers of both parties continuously undertake cross domain transmission and reception hosting.
The sending domain provides source endorsement.
We implement this as the ATP protocol and a prototype system.
ATP locates the receiving entry through the \texttt{local@domain} address and DNS.
It completes custody responsibility transfer with confirmation after verification and saving.
It supports domain source and content integrity verification after message saving through complete envelope signatures and the binding of the sending domain and verification key.

\textbf{Communication continuity and runtime overhead evaluation.} In a controlled workflow involving three administrative domains and driven by LLM agents, all 10 runs of offline and instance replacement variants complete rescheduling processing and hotel booking.
Supporting mechanism experiments verify independent signature verification after message saving.
Under single machine, keep alive connection and memory file system conditions, cross domain ATP adds about 2.7 milliseconds of median latency compared to direct HTTPS.

\section{Background and Problem Definition}
\label{sec:problem}
\subsection{Problem Setup}

This paper focuses on the agent communication problem across different administrative domains.
An administrative domain is an independent management unit that manages local agents, security credentials and access policies.
It can be operated by enterprises, individuals or third party hosting services.
Agents within different administrative domains might be built on heterogeneous underlying models, development frameworks and runtime environments.
They complete business collaboration through intradomain or interdomain message exchange.
Existing research establishes the universal addressing system and message protocol as components of the agent infrastructure \cite{chan2025infrastructure}.

This paper distinguishes four concepts which are logical agent, execution instance, task and network connection.
Among them, the logical agent represents a persistent role participating in cross organization communication.
The execution instance is the underlying process, container or physical device hosting this role at a specific moment.
The same logical agent can be hosted successively by different execution instances during its lifecycle.
A task corresponds to one business interaction with clear completion conditions.
The network connection is responsible for transmitting messages.
In our target scenario, both interacting parties can initiate or receive messages.
The cross domain communication relationship established by both parties often spans multiple tasks and connections.

In actual operation, the execution instance hosting the logical agent may enter a dormant offline state, migrate between different hosts, or run in restricted environments such as behind firewalls and NAT that cannot directly receive inbound public connections.
On the other hand, cross domain communication paths often need to traverse multiple intermediaries and trust boundaries.
The terminal devices and network access points of the participating parties will also switch over time \cite{agentproto-charter}.

\subsection{Communication Requirements}

\textbf{Addressing and discovery.} The communicating party needs to identify the target logical agent and find its receiving entry.
The external address of the logical agent should continue to identify the same communicating party after changes in tasks, connections and execution instances.
The Agent Gateway requirements draft also proposes using a persistent and infrastructure independent Agent ID to identify the communication target.
The gateway queries its current reachability information \cite{liugateway2025v01}.
The discovery mechanism associates the logical address with the corresponding receiving entry.
This makes external organizations not need to track the running location of the instance.
The AID draft publishes the agent service endpoint through DNS TXT records.
The DNS AID draft uses SVCB records to publish connection information and supports agent discovery through organization level indexes \cite{nemethi2026aid,infoblox2026dnsaid}.

\textbf{Continuous reception and delivery.} Messages may arrive when the target execution instance is offline.
They may also be temporarily undeliverable due to network failures.
AgentProto lists path and intermediate node failure recovery and unreachability handling as protocol requirements \cite{agentproto-charter}.
Existing protocols already provide asynchronous interaction mechanisms.
For example, A2A pushes updates through webhooks registered by task \cite{a2a-spec}.
Under the above operating conditions, both cross domain parties need to clarify who accepts and stores messages.
They need to clarify what reception confirmation means.
They also need to clarify who retries after a temporary failure and when the message expires.
The infrastructure receiving the message and the agent completing the business processing have different meanings.
Both parties need to separately judge the delivery result and the business result.

\textbf{Message source verification.} When messages cross administrative domains, the receiver needs to confirm their source and check content integrity.
TLS protects the data transmitted in the connection \cite{rfc8446}.
The WebBotAuth draft defines signature rules for automated HTTP requests and provides a discovery mechanism for verification keys \cite{webbotauth-draft}.
For messages going through temporary storage and delayed delivery, the receiver also needs to verify their source and content after detaching from the original connection.
The source proof should be retained together with the message.
It should cover the content and necessary information of both sending and receiving parties, and retain the verification materials required for reverification.
Delegation and business authorization are handled by corresponding mechanisms \cite{south2025delegation}.
For example, the WIMSE AIMS draft discusses methods of reusing standards like OAuth to implement agent authentication and authorization \cite{rfc6749,wimseaims2026v00}.

\textbf{Cross organization interoperability.} Each administrative domain can independently select internal components.
The communicating parties still need to consistently understand external addresses, reception confirmations and verification rules.
When adopting general security mechanisms, applications also need to clarify their usage methods.
For example, HTTP Message Signatures require applications to specify signature coverage items and key acquisition methods \cite{rfc9421}.
Common external communication rules enable different implementations to interoperate.
Each domain continues to autonomously manage local deployment and access policies.

\subsection{Research Questions}

Endpoints, gateways, queues and signature mechanisms provide technical means to build the above communication capabilities.
Recent related drafts further explore the division of labor and interaction interfaces of the infrastructure.
MACP organizes identity management and capability directories cooperatively through management centers and gateways \cite{macp-draft}.
AGTP supports agents and servers to negotiate methods and endpoints at runtime \cite{agtp-draft}.

At the architecture level, the interdomain collaboration of email and DKIM domain signatures provide mature precedents.
They provide architectural precedents for independent administrative domains to jointly undertake cross domain transmission and source endorsement \cite{rfc5321,rfc6376}.

The core question focused on in this paper is how to organize stable addressing, continuous reception and message source verification in an environment where each organization independently manages local agents and adopts different implementations, in order to support communication relationships crossing changes in tasks, connections and execution instances.
What common rules does this organization method require, and what constraints and costs does it bring?
Section 3 gives the specific design for the domain servers of both parties to continuously participate in cross domain communication.
Section 4 analyzes its design constraints and establishment conditions.

\section{ATP Design}
\label{sec:atp_design}

\subsection{Architecture Overview}

ATP uses each administrative domain to maintain addresses, receive messages and provide domain level source endorsement for logical agents.
Execution instances are responsible for business processing.
Domain servers continuously undertake cross domain transmission and reception hosting.
This division is inspired by the email architecture which organizes addresses by domain, uses servers to cooperate in transmission and saves messages for users.

\begin{figure}[t]
  \centering
  \includegraphics[width=\linewidth]{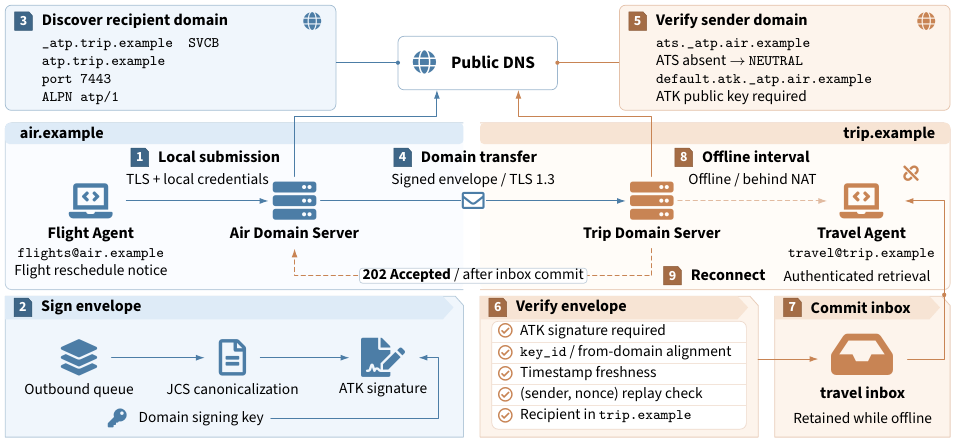}
  \caption[ATP cross-domain message path]{\textbf{ATP cross-domain message path.} \normalfont The sending domain authenticates the local sender, signs the envelope, and discovers the receiving domain through DNS.
  The receiving domain verifies and stores the message for later retrieval by the destination agent.}
  \label{fig:path}
  \Description{The sending agent submits a message to its domain server, which discovers and authenticates the receiving domain and delivers the message to the recipient inbox for later retrieval.}
\end{figure}

As shown in Figure~\ref{fig:path}, a cross-domain message passes through the sending instance, sending domain server, receiving domain server, and receiving instance in sequence.
The sending instance submits the message to its local domain server.
The sending domain server authenticates the local sender, signs the message envelope, and discovers the receiving domain's entry point through DNS.
The receiving domain server verifies the message and stores it in the recipient's inbox.
The receiving instance then retrieves it.
Either party can initiate messages, so the sending and receiving roles change with each message.
Domains follow common rules for discovery, reception acknowledgment, and verification, while independently managing local naming, access policies, and storage deployment.

ATP carries asynchronous messages, requests and responses with deadlines and event subscriptions.
Tasks and dialogues can be organized by upper layer protocols such as A2A \cite{a2a-spec}.
Model context and business status are maintained by the application.
Tasks and authorization materials can be passed with the \texttt{payload}.
Figure~\ref{fig:interaction_flows} summarizes these three interaction patterns; Appendix~\ref{app:protocol} gives their rules.

\subsection{Stable Addressing}

\textbf{ATP manages the address of the logical agent and the running location of the execution instance separately.} Logical agents use the \texttt{local@domain} address.
The domain name determines the naming scope.
The local part is allocated by the domain.
The same logical agent can continue to use the address across tasks and instances.
Each domain names autonomously.
There is no need to jointly rely on a new global registry.

External domains find the receiving service of the domain to which the address belongs through DNS SVCB \cite{rfc9460}.
The receiving domain then locates the local inbox based on the full address.
Instances use the corresponding address after being authenticated by the local domain.
They actively connect to local domain services to send and receive messages.
They do not need to expose public network inbound endpoints.
External domains only need to locate the domain entry and do not need to track instance locations.
When the administrative domain replaces servers or hosting providers, it can also update service records and keep the original agent address.

\subsection{Message Hosting}

ATP uses the receiving domain to save messages for target instances to read later.
This makes cross domain reception not have to wait for instances to go online.
Messages use SMTP style custody responsibility transfer between sending and receiving domains \cite{rfc5321}.
The receiving domain only returns a reception confirmation and assumes subsequent custody responsibility after completing verification and saving.
Before receiving confirmation, the message is kept by the sending domain which is responsible for delivery.
Both sending and receiving parties judge whether the responsibility transfer is complete based on this.
Each domain can still choose different queue and storage implementations.
The business processing results of the agent are expressed through subsequent messages.

Instance offline and receiving domain unreachable affect different stages.
When the instance is offline, the receiving domain continues to accept and save messages.
When the receiving domain is unreachable, the sending domain keeps the message and retries.
Delivery is constrained by the freshness window of cross domain reception.
After the message enters the box, the instance can read it within the inbox retention period.
The retention period is determined by the storage policy of the receiving domain.
Appendix ~\ref{app:protocol} details specific rules of retry, deduplication and failure notification.

ATP continuously retains the domain servers of both parties.
This makes the receiving service not depend on the target instance being online.
SAGA lets agents communicate directly after the initial connection to reduce dependence on the Provider \cite{syros2026saga}.
Continuous mediation increases transmission and server processing overhead.
It also makes communication depend on the availability of domain services.

\subsection{Domain Level Signatures}

ATP uses the sending domain to authenticate local senders and uses the domain private key to provide DKIM style domain level source endorsement for messages \cite{rfc6376}.
The sending domain manages addresses and local credentials.
Execution instances use local credentials to submit messages.
The domain private key is held by the domain service.
We assume DNS responses are trusted, for example verified by DNSSEC \cite{rfc4033}.
We also trust certificate authorities and the operators of sending and receiving domains.

The signature needs to bind the message content and the sending domain at the same time.
ATP signs the complete message object except the signature itself.
This makes the sender address, receiver address and \texttt{payload} protected together.
The public key is published through the Agent Transfer Key (ATK) DNS record.
\textbf{The receiver requires the domain of the public key to match the domain of the sender address.} This prevents attackers from using valid keys of their own domains to impersonate other domains.
Each domain verifies messages according to the same coverage and domain binding requirements.
The form of local credentials is still chosen by each domain.
Storing the signature with the message allows provenance verification after the original connection ends.
A receiving instance can independently reverify domain provenance and content integrity upon retrieval, provided that the original verification key remains available and trusted.

Under the above trust premises, message signatures, sending domain binding and replay prevention checks are used to resist tampering, impersonating other domains and replays by network attackers.
Domain operators hold private keys and can access inboxes.
They belong to the trust boundary.
Domain endorsement confirms the domain source and integrity of messages.
Author identity, delegation relationship and business permissions are verified by corresponding mechanisms \cite{south2025delegation}.
Applications that need to guard against domain operators can add end to end signatures and encryption according to integrity and confidentiality needs.

\subsection{Implementation}

We implemented ATP as a domain server and a command-line client in Go, totaling 5,237 lines of code.
The server handles local message submission, cross-domain delivery, and inbox retrieval through HTTPS interfaces protected by TLS \cite{rfc8446}.
For message signing and verification, it uses JCS to obtain a canonical representation of the unsigned envelope \cite{rfc8785} and Ed25519 to generate and verify signatures over that representation \cite{rfc8032}.
SQLite stores recipient inboxes and the outbound delivery queue.


\section{Combination Scheme Comparison}

This section uses the relevant protocol specifications.
It compares the complete A2A component combination scheme with ATP.
It analyzes the design choices and establishment conditions of both to implement the requirements of Section 2.

\subsection{A2A Combination Scheme}

The combination scheme we select includes the organization resident A2A service, gateways of both parties, persistent queue, OAuth and HTTP Message Signatures (HMS).
Agent Card can be discovered through well known URI, directory or direct configuration \cite{a2a-spec}.
The sending instance submits messages to the local domain gateway through OAuth authorization \cite{rfc6749}.
The gateway signs and delivers them to the receiving organization.
The receiving side verifies and saves them.
The backend instance then reads and processes A2A interactions.
Client metadata can be obtained by authorization servers supporting CIMD (version 02) according to the client ID URL \cite{cimd-draft}.


\subsection{Design Comparison}

\textbf{Addressing and reception.} The combination scheme hides backend instance changes through stable entries and receiving queues.
The association method between logical agents and entries and queues is determined by the selected service and configuration.
ATP uses the \texttt{local@domain} address.
It finds the domain entry through DNS.
The receiving domain then locates the inbox.
Both can receive messages of offline instances through resident services.
The ATP prototype achieves instance replacement by continuing to use addresses, credentials and cursors.
The combination scheme also needs to continue the receiving identity and reading progress.

\textbf{Confirmation and delivery.} The combination scheme needs to select the time point for the gateway to confirm reception.
It needs to connect this with queue saving.
The webhook confirmation of A2A indicates successful reception.
It does not uniformly require persistent saving (Section 4.3.3) \cite{a2a-spec}.
ATP takes verification and saving as the condition for custody responsibility transfer.
The combination scheme can also achieve the same transfer through confirmation after saving.
Both need to handle repeated submissions after confirmation loss.
This makes duplicate judgment consistent with the actual acceptance status.
The queue retention and delivery deadlines of the combination scheme are selected by deployment.
ATP retains the original creation time to retry.
It is limited by the cross domain reception window.
Reading after entering the box is constrained by the inbox retention period.

\textbf{Source verification.} The HMS combination can sign Content Digest and check the body digest.
After saving the signature, content and HTTP information required to rebuild the signature input, it reverifies according to application validity rules \cite{rfc9421,rfc9530}.
WebBotAuth (version 00) provides request signatures.
Its well known key directory can associate publishing domains with keys \cite{webbotauth-draft}.
To provide domain endorsement, the combination scheme needs to associate the verification key with the envelope sending domain.
ATP attaches the signature to the complete envelope.
It obtains the key through DNS.
It requires the domain of the key to match the sending domain.
The difference between the two is the specific selection of signature materials, key discovery and domain binding.
Reverification after saving both depend on complete verification materials and trusted available keys.


\section{Evaluation}
\label{sec:eval}

In this section, We first evaluate the ATP architecture in a complete workflow using a cross organization agent case study.
Next, we verify reception hosting, instance replacement and reverification after saving.
Finally, we measure the continuous mediation overhead and assess the security rules and delivery boundaries.

\subsection{Experimental Environment}

The mechanism, latency, security and unreachability experiments run on a shared Ubuntu 26.04 server (Intel Xeon Gold 6240, 16 cores, 15 GiB memory, Docker 29.1.3, load 1 to 3).
Each domain runs an independent ATP server container.
A BIND9 server provides authoritative resolution.
Certificates are issued by a testing certificate authority.
These experiments use the same build.
Timestamps are taken from the same clock.
The latency experiment places the SQLite database in a memory file system.
It uses WAL and \texttt{synchronous=NORMAL}.
It measures the cost excluding per record disk flushing.
The model of the large model case is \texttt{deepseek-flash}, and the temperature is 0.

\subsection{Cross Organization Agent Case Study}

The case adopts a large model agent workflow among a travel service, an airline and a hotel.
It examines proactive contact after the original task ends, message reading after instance offline or replacement, and cross domain communication when a new organization joins.

The three organizations each have one domain and one agent.
They read and write the organization's data through business tools.
They exchange messages through ATP clients.
The container of the travel agent does not accept inbound connections.
After the ticket booking task is completed, the travel agent goes offline.
The airline subsequently sends a rescheduling notification.
About 360 seconds later, the travel agent goes online to fetch messages, updates the itinerary, and contacts the hotel domain joined by publishing DNS records.
It books a room according to the new arrival date.

We run offline and migration variants respectively, 5 times each.
Prompt debugging before formal operation is not included in the results.
The offline variant is returned by the same instance.
The migration variant uses a new container and network address.
It continues to use the original logical address, credentials and state directory saving reading cursors.
It checks instance replacement within the same hosting domain.
Success judgment is based on five external status checks: the flight booking is consistent with the itinerary, the itinerary is updated to the rescheduled time, the notification is processed only once, the hotel booking date and confirmation number are correct, and the configuration of existing domains remains unchanged before and after the hotel joins.
The results are shown in Table~\ref{tab:case_res}.

\begin{table}[t]
  \centering
  \caption{Travel coordination results (medians over five runs per variant).}
  \ATPCompactTable
  \label{tab:case_res}
  \begin{tabularx}{\columnwidth}{@{}>{\raggedright\arraybackslash}Xrr@{}}
    \toprule
    Variant & Offline & Migration \\
    \midrule
    Success & 5/5 & 5/5 \\
    Total Time (s) & 407.6 & 406.7 \\
    Offline (s) & 360.7 & 361.1 \\
    Model (s) & 35.0 & 34.3 \\
    Business Tool (s) & 0.18 & 0.18 \\
    ATP send/recv (s) & 0.13 / 0.08 & 0.13 / 0.10 \\
    Online to Fetched (s) & 1.66 & 1.63 \\
    \bottomrule
  \end{tabularx}
\end{table}

\begin{figure}[t]
  \ATPCompactFigureCaptions
  \centering
  \begin{subfigure}[t]{\ATPColumnPanelWidth}
    \centering
    \includegraphics[width=\linewidth]{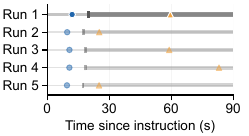}
    \caption{Offline: 0--90\,s.}
    \label{fig:llm_timeline_offline_early}
  \end{subfigure}\hfill
  \begin{subfigure}[t]{\ATPColumnPanelWidth}
    \centering
    \includegraphics[width=\linewidth]{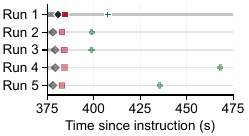}
    \caption{Offline: 375--475\,s.}
    \label{fig:llm_timeline_offline_late}
  \end{subfigure}\par\smallskip
  \begin{subfigure}[t]{\ATPColumnPanelWidth}
    \centering
    \includegraphics[width=\linewidth]{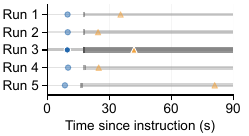}
    \caption{Migration: 0--90\,s.}
    \label{fig:llm_timeline_migration_early}
  \end{subfigure}\hfill
  \begin{subfigure}[t]{\ATPColumnPanelWidth}
    \centering
    \includegraphics[width=\linewidth]{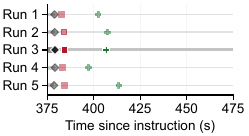}
    \caption{Migration: 375--475\,s.}
    \label{fig:llm_timeline_migration_late}
  \end{subfigure}\par\smallskip
  \includegraphics{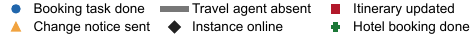}
  \caption{Travel coordination timelines.
  Rows show offline and migration, with five runs each; columns show two time windows from the booking instruction.
  The interval from 90 to 375 seconds is omitted.
  Dark bars mark absence; bold rows identify the runs closest to each variant's median total time.}
  \label{fig:llm_timeline}
  \Description{A two-by-two grid shows all five runs of each variant.
  The top row is offline and the bottom row is migration; the left column spans zero to ninety seconds and the right column spans three hundred seventy-five to four hundred seventy-five seconds.
  A shared legend appears below the panels.}
\end{figure}

All 10 runs of the offline and migration variants complete rescheduling processing and hotel booking.
They pass all checks.
Before and after the hotel joins, the configuration of existing domains remains unchanged.
Figure~\ref{fig:llm_timeline} shows notification sending, instance return and subsequent processing in two time windows for each variant.
The median wait times from notification sending to reading are 323.4 seconds and 346.1 seconds respectively.
The median times from instance return to fetching the notification are 1.66 seconds and 1.63 seconds respectively.
This includes the idle polling every 5 seconds during runtime.
Timing stops when the message fetching returns.
Subsequent itinerary updates and hotel bookings are marked by events in the figure.

\subsection{Mechanism Verification}

We independently check instance offline, instance replacement and source reverification.
Each item runs 5 times.
The offline check verifies quantity and content.
The migration check continues to use the original address, credentials and cursors.
It verifies whether the new instance can read subsequent messages and no longer reads old batches.
We use an independent Python verifier to check messages taken out after saving and their signatures.
The script uses the original verification public key that is still obtainable, and keys are not rotated during operation.
Freshness is checked during cross domain reception.
Reverification does not repeat inbound time window checks.

\begin{table}[t]
  \centering
  \caption{Mechanism checks and observation results, each item runs 5 times.}
  \ATPCompactTable
  \label{tab:machanism_checks}
  \begin{tabularx}{\columnwidth}{@{}>{\raggedright\arraybackslash}p{\ATPCheckLabelWidth}>{\raggedright\arraybackslash}Xc@{}}
    \toprule
    Check & Observation Result for Each Time & Pass \\
    \midrule
    Retrieve after offline (600 seconds) & 20/20 messages received, no missing, duplication or content error & 5/5 \\
    Address unchanged after migration & 20/20 new messages fetched on new IP, old batch not fetched repeatedly & 5/5 \\
    Independent reverification of signature & 20/20 signatures passed verification & 5/5 \\
    Tampering detection & 60/60 modified envelopes rejected & 5/5 \\
    Two agents on one server & 20/20 messages each, no inbox mixing & 5/5 \\
    \bottomrule
  \end{tabularx}
\end{table}

As shown in Table~\ref{tab:machanism_checks}, all checks pass and they support separating message acceptance from instance online status, reading progress continuation after instance replacement, and independent verification of domain source and content integrity after saving.

\subsection{Runtime Overhead}

The latency measurement compares three paths.
Direct HTTPS to the receiving endpoint is used as a reference for measuring incremental overhead.
Intradomain ATP passes through one domain server.
Interdomain ATP passes through two domain servers.
The receiving program fetches messages through long polling.
Receiving programs remain online.
Timing stops when the receiving program gets the message.
All three paths use keep alive HTTPS clients.
The load is 2 and 20 messages per second.
200 and 600 messages are sent each time respectively.
Each runs 3 times.
Message and timing samples are complete.
DNS queries hit the cache after warmup.

\begin{figure}[t]
  \ATPCompactFigureCaptions
  \centering
  \begin{subfigure}[t]{\ATPColumnPanelWidth}
    \centering
    \includegraphics[width=\linewidth]{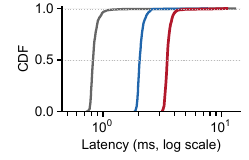}
    \caption{2 msg/s ($n=600$).}
    \label{fig:latency_cdf_low}
  \end{subfigure}\hfill
  \begin{subfigure}[t]{\ATPColumnPanelWidth}
    \centering
    \includegraphics[width=\linewidth]{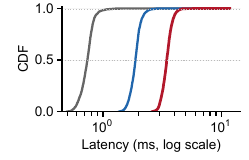}
    \caption{20 msg/s ($n=1{,}800$).}
    \label{fig:latency_cdf_medium}
  \end{subfigure}\par\smallskip
  \includegraphics{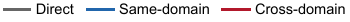}
  \caption{End-to-end latency CDFs at two loads, with shared logarithmic axes.
  Dotted guides mark p50 and p99.}
  \label{fig:latency_cdf}
  \Description{Two side-by-side CDF panels compare direct HTTPS, same-domain ATP, and cross-domain ATP at two and twenty messages per second.
  Both panels use identical logarithmic latency limits and share a path-color legend below.}
\end{figure}

Interdomain ATP adds about 2.7 milliseconds of median latency compared to direct HTTPS.
It adds about 1.5 milliseconds compared to intradomain ATP.
As shown in Figure~\ref{fig:latency_cdf}, the median latencies of interdomain ATP under the two loads are 3.48 and 3.45 milliseconds respectively.
The p99 are 4.72 and 4.47 milliseconds respectively.
The latency distributions of the two load levels are close.

\begin{figure}[t]
  \ATPCompactFigureCaptions
  \centering
  \begin{subfigure}[t]{\ATPColumnPanelWidth}
    \centering
    \includegraphics[width=\linewidth]{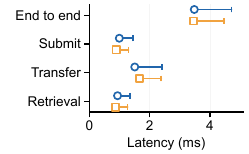}
    \caption{Path segments.}
    \label{fig:cross_domain_segments}
  \end{subfigure}\hfill
  \begin{subfigure}[t]{\ATPColumnPanelWidth}
    \centering
    \includegraphics[width=\linewidth]{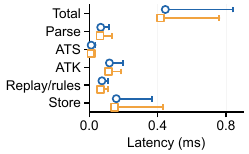}
    \caption{Receiver stages.}
    \label{fig:receiver_stages}
  \end{subfigure}\par\smallskip
  \includegraphics{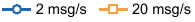}
  \caption{Runtime latency breakdown.
  Open markers denote p50; segment ends denote p99.}
  \label{fig:runtime_cost}
  \Description{Two side-by-side panels show the median-to-ninety-ninth-percentile latency intervals for cross-domain path segments and receiving-server processing stages.
  A shared legend below identifies the two traffic loads.}
\end{figure}

Interdomain delivery is the most time consuming segment in the cross domain path.
Its median time consumption is 1.51 and 1.67 milliseconds.
Figure~\ref{fig:runtime_cost} breaks down the path into submission, interdomain delivery and local retrieval (panel~\subref{fig:cross_domain_segments}), alongside receiving-domain processing (panel~\subref{fig:receiver_stages}).
Submission is counted until the sending domain writes.
Interdomain delivery is counted until the receiving domain writes to the inbox.
It includes sending domain signature.
Local retrieval is counted until the receiving program gets the message.

Inbox writing and signature verification occupy the main processing time of the receiving end.
The median total processing of the receiving end is 0.445 and 0.418 milliseconds.
Figure~\ref{fig:receiver_stages} decomposes the time consumption of parsing, ATS source address filtering, ATK signature verification, replay prevention and rule checking (including recipient query), and inbox writing.

These results quantify the incremental overhead of the ATP path relative to direct HTTPS under single machine, keep alive connection and memory file system conditions.
Wide area deployment is also affected by network delay and storage configuration.
When the database is moved to a shared disk, the median is similar.
But I/O interference from other users raises p99 to the second level.
When domain services undertake saving responsibilities, their storage selection also affects tail latency.

\subsection{Security Testing}

The eight types of malicious inputs listed in Table~\ref{tab:security_checks} are used to verify source verification and admission rules.
Each test case runs 3 times.
We check the returned status codes and error codes.
We confirm messages do not enter the inbox.
Each test case only violates the one rule under test.
The sending domain publishes the public key but omits the ATS record, ensuring the signature mechanism evaluates the forged signature and other domain key cases.
For the other domain key case, the signature is valid but its domain mismatches the sending domain.
This verifies the domain binding requirements in Section 3.

\begin{table}[t]
  \centering
  \caption{Security checks.
  Each test case runs 3 times, all are rejected by expected mechanisms, messages do not enter the inbox.}
  \ATPCompactTable
  \label{tab:security_checks}
  \begin{tabularx}{\columnwidth}{@{}>{\raggedright\arraybackslash}p{\ATPCheckLabelWidth}>{\raggedright\arraybackslash}Xc@{}}
    \toprule
    Test Case & Rejected by Which Check (Status Code) & Pass \\
    \midrule
    Forged signature & Signature check (403) & 3/3 \\
    Tampered content & Signature check (403) & 3/3 \\
    Use other domain key & Domain consistency (403) & 3/3 \\
    Unauthorized source address & ATS policy (403) & 3/3 \\
    Forged source request header & ATS policy (403) & 3/3 \\
    Replay & Replay prevention cache (400) & 3/3 \\
    Expired timestamp & Freshness window (400) & 3/3 \\
    Future timestamp & Freshness window (400) & 3/3 \\
    \bottomrule
  \end{tabularx}
\end{table}

\subsection{Delivery When Receiving Domain is Unreachable}

\begin{figure*}[t]
  \centering
  \begin{subfigure}[t]{\ATPWidePanelWidth}
    \centering
    \includegraphics[width=\linewidth]{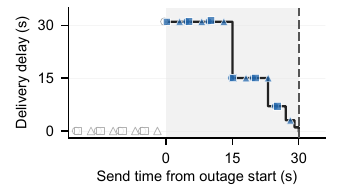}
    \caption{30\,s outage ($n=30$).}
    \label{fig:outage_30}
  \end{subfigure}\hfill
  \begin{subfigure}[t]{\ATPWidePanelWidth}
    \centering
    \includegraphics[width=\linewidth]{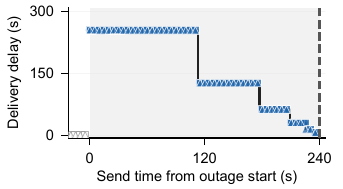}
    \caption{240\,s outage ($n=156$).}
    \label{fig:outage_240}
  \end{subfigure}\hfill
  \begin{subfigure}[t]{\ATPWidePanelWidth}
    \centering
    \includegraphics[width=\linewidth]{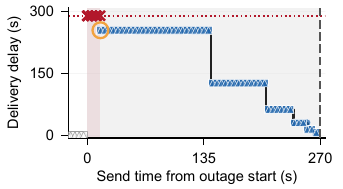}
    \caption{270\,s outage ($n=174$).}
    \label{fig:outage_270}
  \end{subfigure}\par\smallskip
  \includegraphics{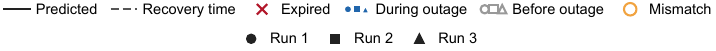}
  \caption{Delivery under receiver outages, with three runs per duration.
  The x-axis is send time relative to outage start.
  The staircase predicts delivery delay; crosses mark expired messages at the 290\,s cutoff.
  The orange circle marks the single prediction mismatch.
  Shading indicates the outage.}
  \label{fig:outage}
  \Description{Three equal-height panels in one row across both columns show all messages during thirty-, two-hundred-forty-, and two-hundred-seventy-second outages.
  Predicted delay, recovery, expired messages, repeated runs, send timing, and the single mismatch are explained by a shared legend below the panels.}
\end{figure*}

This experiment checks whether the sending domain can complete delivery within the receiving window after the receiving domain recovers.
Under the condition that connections are immediately rejected, default delivery attempts occur at approximately message ages of 0, 1, 3, 7 ... 127 and 255 seconds.
The next attempt will exceed the upper age limit of 290 seconds.
We set unreachability durations of 30, 240 and 270 seconds accordingly.
This covers short interruptions and recovery before and after the last valid attempt of early messages.
The receiving server keeps running and replies TCP RST to new connections during this period.
The sender sends one message every 5 seconds.
Each group runs 3 times.
Figure~\ref{fig:outage} compares the observed outcomes with the retry schedule for all three outage durations.

All messages in the 30 second and 240 second groups are delivered.
The 270 second group has 3 to 4 expired messages each time.
All are sent within 15 seconds after unreachability begins.
The remaining 54 to 55 messages are delivered.
No duplicated, other permanent failures or messages still waiting to be delivered are observed.
Actual attempt time consumption shifts the predicted boundary.
A message sent 15.03 seconds after unreachability begins has its last attempt postponed until after recovery due to previous time consumption.
Thus it differs from the expired result predicted by the ideal schedule.
After recovery, messages still need to wait for the next scheduled retry.
The median wait time of the 240 second group is 36 seconds, maximum 125 seconds.

This result reflects the two deadlines distinguished in Section 4.
Interdomain delivery is constrained by the receiving window.
Local reading after messages enter the inbox is constrained by the inbox retention policy.
In the aforementioned experiments, messages saved by the receiving domain can still be retrieved 360 seconds or 600 seconds after the instance goes offline.

\section{Related Work}

\textbf{Agentic web and shared infrastructure.} Research on Agentic Web, Web of Agents and Internet of Agents discusses agent ecosystems around Web services, autonomous collaboration and network interconnection \cite{yang2025agenticweb,petrova2025webofagents,wang2025ioafundamentals}.
Chan et al. define agent infrastructure as technical systems and shared protocols independent of agents.
They use external mechanisms to organize interactions among heterogeneous participants \cite{chan2025infrastructure}.
Sander et al. compare protocol interoperability mechanisms from dimensions such as communication objects, payloads, states and discovery methods \cite{sander2026taxonomy}.
We apply the perspective of shared infrastructure to cross domain communication.
We study the design where domain services maintain addresses, host messages and provide domain level source endorsement for logical agents.

\textbf{Agent communication protocols and access architectures.} Application interaction protocols and communication infrastructure assume responsibilities at different levels.
A2A defines message, task and context interactions between clients and agent servers.
It supports polling, streaming updates and task pushing.
It requires servers to authenticate requests \cite{a2a-spec}.
MCP organizes interactions between clients and servers through tool, resource and prompt interfaces \cite{mcp-spec}.
The server side can be hosted by persistent services.
It hides backend instance changes through gateways and queues.
Aiming at model differences, LDP incorporates model identity information, payload negotiation and managed sessions into delegation protocols for task routing and communication \cite{prakash2026ldp}.

Identity, discovery and access services also have different levels of participation in message transmission.
SAGA uses Providers to manage registration, discovery and user defined contact permissions.
Agents communicate directly after obtaining contact information and access materials.
Providers do not forward subsequent messages.
The system assumes endpoints have public network routable addresses \cite{syros2026saga}.
ANP combines DID identity authentication, agent description and discovery, and protocol negotiation.
It supports interconnection in open networks \cite{anp-whitepaper}.
MACP organizes identity and credential management, capability registration, cross gateway capability synchronization and collaborative request processing into management centers and gateways \cite{macp-draft}.
The managed gateway of AgentCore can uniformly execute entry authentication and policies outside the agent running environment \cite{aws-agentcore-runtime}.

\textbf{Long term interaction, recovery and message hosting.} Long term communication involves the continuation of dialogs.
It also involves saving and subsequent delivery of messages during interruptions.
Related individual drafts distinguish sessions from underlying connections.
They discuss conditions for maintaining continuity after endpoint or running environment changes \cite{feng2026sessions}.
The PACE draft supports session pause and recovery through MOQT publish subscribe and relay caching.
It fetches missed objects from caches \cite{nandakumar2026pace}.
The platform analysis of Agentverse by Dey et al. describes the mechanism of asynchronous mailboxes temporarily storing messages when target agents are offline \cite{dey2026agentverse}.
These works organize recovery and hosting from sessions, relays and platform services respectively.
ATP targets message delivery between independent administrative domains.
It separates the custody responsibility of receiving domains from the business processing of execution instances.
It specifies interdomain reception confirmation, sending side retry and message verification rules.
Upper layer protocols and applications continue to handle dialog association and business state recovery.

\textbf{Identity, delegation and message security.} The individual draft by Bu distinguishes logical agents, execution instances and human or organization authorized subjects.
It discusses current instance authentication, binding of instances and logical identities, and verification of historical continuity respectively \cite{bu2026principalbinding}.
South et al. discuss verifiable delegation subjects and permission scopes \cite{south2025position}.
AIP uses credential chains supporting permission attenuation.
It provides delegation verification across MCP, A2A and HTTP \cite{prakash2026aip}.
BlockA2A combines DID, signed messages and on chain records to support identity and interaction verification \cite{zou2025blocka2a}.

\textbf{Internet communication precedents.} Interdomain message hosting, stable addressing and savable source credentials provide the mechanism foundation for ATP.
SMTP uses domain servers to cooperatively receive and forward emails.
DKIM associates signature domains with single messages \cite{rfc5321,rfc6376}.
SIP and Mobile IP separate stable identifiers and current locations at the application session and network levels respectively \cite{rfc3261,rfc6275}.
XMPP provides message routing organized by domain.
Matrix synchronizes room events with source signatures among federated servers.
ActivityPub defines server to server activity delivery targeting actor inboxes \cite{rfc6120,matrix-s2s,activitypub}.
ATP targets peer to peer messages of cross organization logical agents.
It reuses interdomain cooperation, message hosting and message signature mechanisms.
It incorporates logical addresses, DNS entry discovery, reception confirmation after saving and sending domain binding into a distributed communication protocol.

\section{Conclusion}
\label{sec:conclusion}

We propose Agent Transfer Protocol (ATP), using domain names as administrative boundaries for providing source endorsement and continuous message reception for cross organization agent-to-agent communication.
ATP servers maintain stable addresses and receive and save messages when target execution instances are offline, providing an implementable communication architecture for agent interactions crossing task and instance changes.

\clearpage

\FloatBarrier
\bibliographystyle{\ATPBibliographyStyle}
\bibliography{references}

@techreport{rfc5321,
  author      = {John Klensin},
  title       = {Simple Mail Transfer Protocol},
  type        = {RFC},
  number      = {5321},
  institution = {IETF},
  year        = {2008},
  month       = oct,
  doi         = {10.17487/RFC5321},
}

@techreport{rfc7208,
  author      = {Scott Kitterman},
  title       = {Sender Policy Framework ({SPF}) for Authorizing Use of Domains in Email, Version 1},
  type        = {RFC},
  number      = {7208},
  institution = {IETF},
  year        = {2014},
  month       = apr,
  doi         = {10.17487/RFC7208},
}

@techreport{rfc6376,
  author      = {Dave Crocker and Tony Hansen and Murray Kucherawy},
  title       = {{DomainKeys} Identified Mail ({DKIM}) Signatures},
  type        = {RFC},
  number      = {6376},
  institution = {IETF},
  year        = {2011},
  month       = sep,
  doi         = {10.17487/RFC6376},
}

@techreport{rfc6120,
  author      = {Peter Saint-Andre},
  title       = {Extensible Messaging and Presence Protocol ({XMPP}): Core},
  type        = {RFC},
  number      = {6120},
  institution = {IETF},
  year        = {2011},
  month       = mar,
  doi         = {10.17487/RFC6120},
}

@misc{matrix-s2s,
  author       = {{The Matrix.org Foundation}},
  title        = {Matrix Specification: Server-Server {API}, Version v1.19},
  year         = {2026},
  url          = {https://spec.matrix.org/latest/server-server-api/},
}

@misc{activitypub,
  author       = {Christine Lemmer-Webber and Jessica Tallon and Erin Shepherd and Amy Guy and Evan Prodromou},
  title        = {{ActivityPub}},
  howpublished = {W3C Recommendation},
  year         = {2018},
  month        = jan,
  url          = {https://www.w3.org/TR/activitypub/},
}

@misc{agentproto-charter,
  author       = {{IETF}},
  title        = {Charter for the Agent Communication Protocols (agentproto) Working Group},
  howpublished = {IETF working group charter, charter-ietf-agentproto-01},
  year         = {2026},
  month        = oct,
  url          = {https://datatracker.ietf.org/doc/charter-ietf-agentproto/01/},
}

@misc{aws-agentcore-runtime,
  author       = {{Amazon Web Services}},
  title        = {Authenticate and Authorize with Inbound Auth and Outbound Auth},
  howpublished = {Amazon Bedrock AgentCore Developer Guide},
  year         = {2026},
  url          = {https://docs.aws.amazon.com/bedrock-agentcore/latest/devguide/runtime-oauth.html},
}

@misc{a2a-spec,
  author       = {{A2A Project}},
  title        = {Agent2Agent ({A2A}) Protocol Specification, Version 1.0.0},
  year         = {2026},
  url          = {https://a2a-protocol.org/v1.0.0/specification/},
}

@misc{cimd-draft,
  author       = {Aaron Parecki and Emelia Smith},
  title        = {{OAuth} Client {ID} Metadata Document},
  howpublished = {IETF Internet-Draft draft-ietf-oauth-client-id-metadata-document-02},
  year         = {2026},
  month        = jul,
  url          = {https://datatracker.ietf.org/doc/draft-ietf-oauth-client-id-metadata-document/},
}

@techreport{webbotauth-draft,
  author      = {Thibault Meunier and Sandor Major},
  title       = {{HTTP} Message Signatures for automated traffic},
  type        = {Internet-Draft},
  number      = {draft-ietf-webbotauth-httpsig-protocol-00},
  institution = {Internet Engineering Task Force},
  year        = {2026},
  month       = sep,
  day         = {1},
  url         = {https://datatracker.ietf.org/doc/draft-ietf-webbotauth-httpsig-protocol/00/},
  note        = {Work in progress}
}

@techreport{rfc9421,
  author      = {Annabelle Backman and Justin Richer and Manu Sporny},
  title       = {{HTTP} Message Signatures},
  type        = {RFC},
  number      = {9421},
  institution = {IETF},
  year        = {2024},
  month       = feb,
  doi         = {10.17487/RFC9421},
}

@techreport{rfc6749,
  author      = {Dick Hardt},
  title       = {The {OAuth} 2.0 Authorization Framework},
  type        = {RFC},
  number      = {6749},
  institution = {IETF},
  year        = {2012},
  month       = oct,
  doi         = {10.17487/RFC6749},
}

@techreport{rfc9530,
  author      = {Roberto Polli and Lucas Pardue},
  title       = {Digest Fields},
  type        = {RFC},
  number      = {9530},
  institution = {IETF},
  year        = {2024},
  month       = feb,
  doi         = {10.17487/RFC9530},
}

@techreport{macp-draft,
  author      = {Xueting Li and Bing Liu and Jun Liu and Chenguang Du and Lianhua Zhang},
  title       = {Multi-agent Collaboration Protocol Suites Architecture},
  type        = {Internet-Draft},
  number      = {draft-li-dmsc-macp-06},
  institution = {Internet Engineering Task Force},
  year        = {2026},
  month       = jul,
  day         = {20},
  url         = {https://datatracker.ietf.org/doc/draft-li-dmsc-macp/06/},
  note        = {Work in progress}
}

@techreport{agtp-draft,
  author      = {Chris Hood},
  title       = {Agent Transfer Protocol ({AGTP})},
  type        = {Internet-Draft},
  number      = {draft-hood-independent-agtp-09},
  institution = {Internet Engineering Task Force},
  year        = {2026},
  month       = jun,
  day         = {28},
  url         = {https://datatracker.ietf.org/doc/draft-hood-independent-agtp/09/},
  note        = {Work in progress}
}

@inproceedings{syros2026saga,
  author    = {Georgios Syros and Anshuman Suri and Jacob Ginesin and Cristina Nita-Rotaru and Alina Oprea},
  title     = {{SAGA}: A Security Architecture for Governing {AI} Agentic Systems},
  booktitle = {Proceedings of the Network and Distributed System Security Symposium (NDSS)},
  year      = {2026},
  doi       = {10.14722/ndss.2026.230869},
}

@techreport{shavit2023practices,
  author      = {Yonadav Shavit and Sandhini Agarwal and Miles Brundage and Steven Adler and Cullen O'Keefe and Rosie Campbell and Teddy Lee and Pamela Mishkin and Tyna Eloundou and Alan Hickey and Katarina Slama and Lama Ahmad and Paul McMillan and Alex Beutel and Alexandre Passos and David G. Robinson},
  title       = {Practices for Governing Agentic {AI} Systems},
  institution = {OpenAI},
  type        = {White paper},
  year        = {2023},
  url         = {https://cdn.openai.com/papers/practices-for-governing-agentic-ai-systems.pdf},
}

@article{chan2025infrastructure,
  author  = {Alan Chan and Kevin Wei and Sihao Huang and Nitarshan Rajkumar and Elija Perrier and Seth Lazar and Gillian K. Hadfield and Markus Anderljung},
  title   = {Infrastructure for {AI} Agents},
  journal = {Transactions on Machine Learning Research (TMLR)},
  year    = {2025},
  url     = {https://arxiv.org/abs/2501.10114v3},
}

@article{south2025delegation,
  author  = {Tobin South and Samuele Marro and Thomas Hardjono and Robert Mahari and Cedric Deslandes Whitney and Dazza Greenwood and Alan Chan and Alex Pentland},
  title   = {Authenticated Delegation and Authorized {AI} Agents},
  journal = {arXiv preprint arXiv:2501.09674},
  year    = {2025},
}

@article{yang2025agenticweb,
  author  = {Yingxuan Yang and Mulei Ma and Yuxuan Huang and Huacan Chai and Chenyu Gong and Haoran Geng and Yuanjian Zhou and Ying Wen and others},
  title   = {Agentic Web: Weaving the Next Web with {AI} Agents},
  journal = {arXiv preprint arXiv:2507.21206},
  year    = {2025},
}

@techreport{rfc9460,
  author      = {Benjamin Schwartz and Mike Bishop and Erik Nygren},
  title       = {Service Binding and Parameter Specification via the {DNS} ({SVCB} and {HTTPS} Resource Records)},
  type        = {RFC},
  number      = {9460},
  institution = {IETF},
  year        = {2023},
  month       = nov,
  doi         = {10.17487/RFC9460},
}

@techreport{rfc4033,
  author      = {Roy Arends and Rob Austein and Matt Larson and Dan Massey and Scott Rose},
  title       = {{DNS} Security Introduction and Requirements},
  type        = {RFC},
  number      = {4033},
  institution = {IETF},
  year        = {2005},
  month       = mar,
  doi         = {10.17487/RFC4033},
}

@techreport{rfc8446,
  author      = {Eric Rescorla},
  title       = {The Transport Layer Security ({TLS}) Protocol Version 1.3},
  type        = {RFC},
  number      = {8446},
  institution = {IETF},
  year        = {2018},
  month       = aug,
  doi         = {10.17487/RFC8446},
}

@techreport{rfc8032,
  author      = {Simon Josefsson and Ilari Liusvaara},
  title       = {Edwards-Curve Digital Signature Algorithm ({EdDSA})},
  type        = {RFC},
  number      = {8032},
  institution = {IETF},
  year        = {2017},
  month       = jan,
  doi         = {10.17487/RFC8032},
}

@techreport{rfc8785,
  author      = {Anders Rundgren and Bret Jordan and Samuel Erdtman},
  title       = {{JSON} Canonicalization Scheme ({JCS})},
  type        = {RFC},
  number      = {8785},
  institution = {IETF},
  year        = {2020},
  month       = jun,
  doi         = {10.17487/RFC8785},
}

@techreport{rfc3261,
  author      = {Jonathan Rosenberg and Henning Schulzrinne and Gonzalo Camarillo and Alan Johnston and Jon Peterson and Robert Sparks and Mark Handley and Eve Schooler},
  title       = {{SIP}: Session Initiation Protocol},
  type        = {RFC},
  number      = {3261},
  institution = {IETF},
  year        = {2002},
  month       = jul,
  doi         = {10.17487/RFC3261},
}

@techreport{rfc6275,
  author      = {Charles E. Perkins and David B. Johnson and Jari Arkko},
  title       = {Mobility Support in {IPv6}},
  type        = {RFC},
  number      = {6275},
  institution = {IETF},
  year        = {2011},
  month       = jul,
  doi         = {10.17487/RFC6275},
}

@techreport{nemethi2026aid,
  author      = {Balazs Nemethi},
  title       = {Agent Identity and Discovery ({AID})},
  type        = {Internet-Draft},
  number      = {draft-nemethi-aid-agent-identity-discovery-00},
  institution = {Internet Engineering Task Force},
  year        = {2026},
  month       = mar,
  url         = {https://datatracker.ietf.org/doc/draft-nemethi-aid-agent-identity-discovery/00/},
  note        = {Work in progress}
}

@techreport{infoblox2026dnsaid,
  author      = {Jim Mozley and Nic Williams and Behcet Sarikaya and Roland Schott and Jeffrey Damick},
  title       = {{DNS} for {AI} Discovery},
  type        = {Internet-Draft},
  number      = {draft-mozleywilliams-dnsop-dnsaid-02},
  institution = {Internet Engineering Task Force},
  year        = {2026},
  month       = may,
  day         = {27},
  url         = {https://datatracker.ietf.org/doc/draft-mozleywilliams-dnsop-dnsaid/02/},
  note        = {Work in progress}
}

@techreport{wimseaims2026v00,
  author      = {Pieter Kasselman and Jeff Lombardo and Yaroslav Rosomakho and Brian Campbell and Nick Steele and Aaron Parecki},
  title       = {{AI} Identity Management System},
  type        = {Internet-Draft},
  number      = {draft-ietf-wimse-aims-00},
  institution = {Internet Engineering Task Force},
  year        = {2026},
  month       = sep,
  day         = {15},
  url         = {https://datatracker.ietf.org/doc/draft-ietf-wimse-aims/00/},
  note        = {Work in progress}
}

@techreport{liugateway2025v01,
  author      = {Bing Liu and Nan Geng and Xiaotong Shang and Qiangzhou Gao and Zhenbin Li and Jing Gao},
  title       = {Requirements for Agent Gateway},
  type        = {Internet-Draft},
  number      = {draft-liu-rtgwg-agent-gateway-requirements-01},
  institution = {Internet Engineering Task Force},
  year        = {2025},
  month       = nov,
  day         = {27},
  url         = {https://datatracker.ietf.org/doc/draft-liu-rtgwg-agent-gateway-requirements/01/},
  note        = {Work in progress; expired Internet-Draft}
}

@article{wang2025ioafundamentals,
  author       = {Wang, Yuntao and Guo, Shaolong and Pan, Yanghe and Su, Zhou and Chen, Fahao and Luan, Tom H. and Li, Peng and Kang, Jiawen and Niyato, Dusit},
  title        = {{Internet of Agents: Fundamentals, Applications, and Challenges}},
  journal      = {arXiv preprint arXiv:2505.07176},
  year         = {2025},
  url          = {https://arxiv.org/abs/2505.07176v2},
}

@article{petrova2025webofagents,
  author       = {Petrova, Tatiana and Bliznioukov, Boris and Puzikov, Aleksandr and State, Radu},
  title        = {{From Multi-Agent Systems and the Semantic Web to Agentic AI: A Unified Narrative of the Web of Agents}},
  journal      = {arXiv preprint arXiv:2507.10644},
  year         = {2025},
  url          = {https://arxiv.org/abs/2507.10644v4},
}

@article{zou2025blocka2a,
  author       = {Zou, Zhenhua and Liu, Zhuotao and Zhao, Lepeng and Zhan, Qiuyang},
  title        = {{BlockA2A: Towards Secure and Verifiable Agent-to-Agent Interoperability}},
  journal      = {arXiv preprint arXiv:2508.01332},
  year         = {2025},
  url          = {https://arxiv.org/abs/2508.01332v3},
}

@article{prakash2026ldp,
  author       = {Prakash, Sunil},
  title        = {{LDP: An Identity-Aware Protocol for Multi-Agent LLM Systems}},
  journal      = {arXiv preprint arXiv:2603.08852},
  year         = {2026},
  url          = {https://arxiv.org/abs/2603.08852v1},
}

@article{prakash2026aip,
  author       = {Prakash, Sunil},
  title        = {{AIP: Agent Identity Protocol for Verifiable Delegation Across MCP and A2A}},
  journal      = {arXiv preprint arXiv:2603.24775},
  year         = {2026},
  url          = {https://arxiv.org/abs/2603.24775v1},
}

@article{sander2026taxonomy,
  author       = {Sander, Linus and Gidey, Habtom Kahsay and Lenz, Alexander and Knoll, Alois},
  title        = {{A Technical Taxonomy of LLM Agent Communication Protocols}},
  journal      = {arXiv preprint arXiv:2606.19135},
  year         = {2026},
  url          = {https://arxiv.org/abs/2606.19135v1},
}

@article{dey2026agentverse,
  author       = {Dey, Robin and Viradecha, Panyanon},
  title        = {{Infrastructure for the Agentic Web: Gap Analysis and Architecture from the Agentverse Platform}},
  journal      = {arXiv preprint arXiv:2606.20570},
  year         = {2026},
  url          = {https://arxiv.org/abs/2606.20570v1},
}

@inproceedings{south2025position,
  author       = {Tobin South and Samuele Marro and Thomas Hardjono and Robert Mahari and Cedric Deslandes Whitney and Alan Chan and Alex Pentland},
  title        = {Position: {AI} Agents Need Authenticated Delegation},
  booktitle    = {Proceedings of the 42nd International Conference on Machine Learning},
  series       = {Proceedings of Machine Learning Research},
  volume       = {267},
  pages        = {82211--82231},
  publisher    = {PMLR},
  year         = {2025},
  url          = {https://proceedings.mlr.press/v267/south25a.html},
}

@techreport{feng2026sessions,
  author       = {Chong Feng},
  title        = {Requirements for Agent Session Establishment, Capability Negotiation, and Sessionless Interaction},
  type         = {Internet-Draft},
  number       = {draft-feng-agentproto-session-requirements-02},
  institution  = {Internet Engineering Task Force},
  year         = {2026},
  url          = {https://www.ietf.org/archive/id/draft-feng-agentproto-session-requirements-02.html},
  note         = {Work in progress},
}

@techreport{bu2026principalbinding,
  author       = {Songbo Bu},
  title        = {Security Principal and Verifier Binding for Agent Communication Protocols},
  type         = {Internet-Draft},
  number       = {draft-bu-agentproto-security-principal-binding-08},
  institution  = {Internet Engineering Task Force},
  year         = {2026},
  month        = oct,
  day          = {14},
  url          = {https://www.ietf.org/archive/id/draft-bu-agentproto-security-principal-binding-08.html},
  note         = {Work in progress},
}

@techreport{nandakumar2026pace,
  author       = {Suhas Nandakumar and Cullen Jennings},
  title        = {{PACE}: Protocol for Agent Communication Exchange},
  type         = {Internet-Draft},
  number       = {draft-nandakumar-agentproto-moq-pace-00},
  institution  = {Internet Engineering Task Force},
  year         = {2026},
  url          = {https://www.ietf.org/archive/id/draft-nandakumar-agentproto-moq-pace-00.html},
  note         = {Work in progress},
}

@misc{anp-whitepaper,
  author       = {{AI Agent Protocol Community Group}},
  title        = {Agent Network Protocol White Paper},
  howpublished = {Community white paper, accessed 2026-10-26},
  year         = {2026},
  url          = {https://w3c-cg.github.io/ai-agent-protocol/},
}

@misc{mcp-spec,
  author       = {{Model Context Protocol}},
  title        = {Model Context Protocol Specification, Revision 2026-07-28},
  year         = {2026},
  url          = {https://modelcontextprotocol.io/specification/2026-07-28},
}

\appendix
\section{Protocol Details}
\label{app:protocol}

This appendix gives the protocol details.

\begin{figure*}[t]
  \centering
  \begin{subfigure}[t]{\ATPMessagePanelWidth}
    \centering
    \includegraphics[height=\ATPFlowHeight]{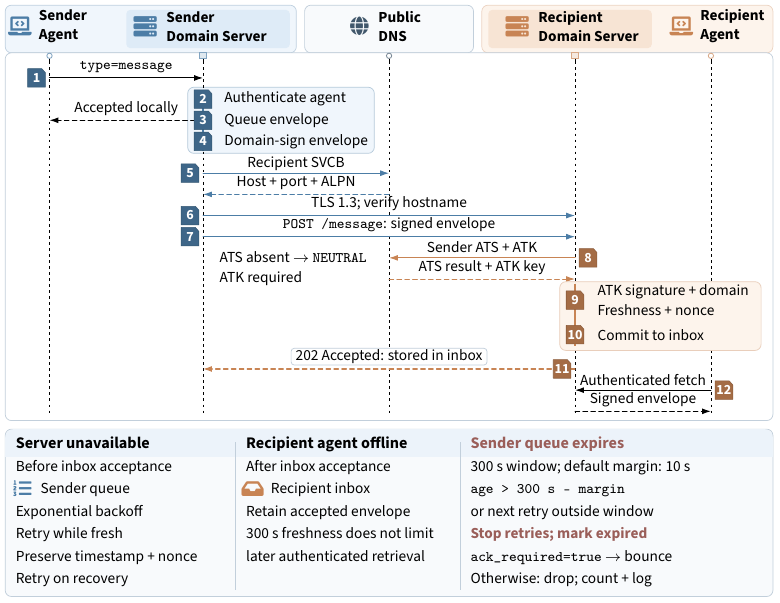}
    \caption{Message delivery.}
    \label{fig:message-delivery}
  \end{subfigure}\hfill
  \begin{subfigure}[t]{\ATPRequestPanelWidth}
    \centering
    \includegraphics[height=\ATPFlowHeight]{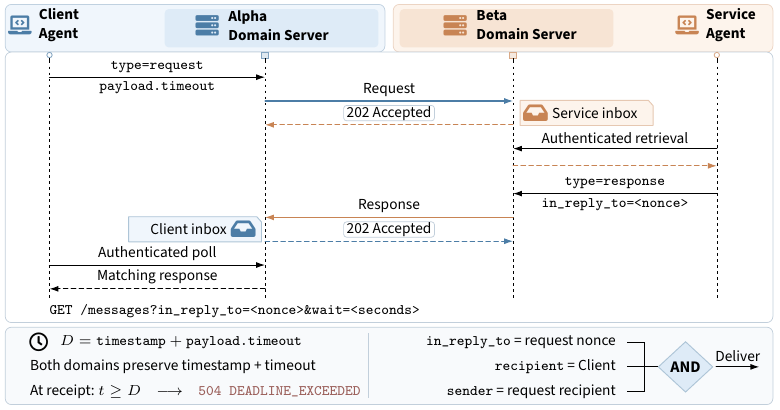}
    \caption{Request/response.}
    \label{fig:request-response}
  \end{subfigure}\hfill
  \begin{subfigure}[t]{\ATPSubscriptionPanelWidth}
    \centering
    \includegraphics[height=\ATPFlowHeight]{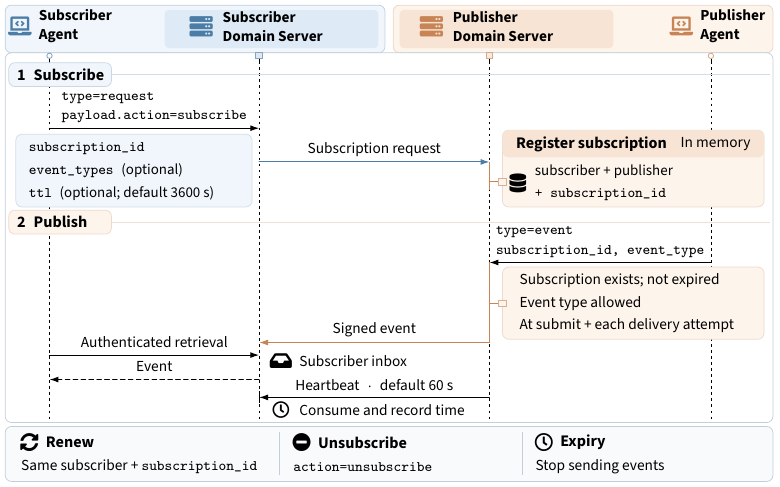}
    \caption{Event subscription.}
    \label{fig:publish-subscribe}
  \end{subfigure}
  \caption{ATP interaction flows: message delivery and custody transfer, request/response deadlines and matching, and the subscription lifecycle.}
  \label{fig:interaction_flows}
  \Description{Three equal-height diagrams appear side by side across both columns.
  They illustrate asynchronous message delivery, request and response with deadline checks, and event subscription with renewal, cancellation, and expiry.}
\end{figure*}

\para{Addressing and discovery.} The sending server queries the SVCB record of \texttt{\_atp.<domain>}~\cite{rfc9460} for the recipient's domain and obtains the host name and port (default 7443) of the receiving server:
\begin{verbatim}
_atp.air.example. IN SVCB 1 atp.air.example. (
    port=7443 alpn="atp/1" )
\end{verbatim}
If the record does not exist, the sender marks the message as a permanent failure.
With several records, the sender tries them in priority order but moves to the next server only when a connection could not be established.
Once a request may have been sent, it is not sent to another server, so that two servers never accept the same message.
Each server lists its supported interaction types at \texttt{/.well-known/atp/v1/capabilities}.
A domain can let a hosting provider run its server by pointing the SVCB record at the provider, which presents a certificate valid for that host name.
The SVCB record describes only the domain's entry point and does not reveal which participants exist.
Servers connect over TLS~\cite{rfc8446}, and the sender verifies the peer certificate against the host name in the SVCB record.
A receiving domain accepts only messages for its own recipients, returning 403 when the recipient belongs to another domain and 404 when the recipient does not exist.

\para{Envelope.} All messages share one JSON envelope.
The required fields are \texttt{from}, \texttt{to}, \texttt{timestamp} (integer seconds), \texttt{nonce}, and \texttt{type}.
\texttt{in\_reply\_to} links a response to its request, and \texttt{payload} is defined by the application.
A rescheduling notice looks as follows:
\begin{verbatim}
{"from": "flights@air.example",
 "to": "travel@trip.example",
 "timestamp": 1792500000,
 "nonce": "n-7f3a9c",
 "type": "message",
 "payload": {"booking": "AX7Q2L",
             "event": "schedule_change",
             "new_departure": "2026-11-03T09:40:00Z"},
 "signature": {
   "key_id": "default.atk._atp.air.example",
   "algorithm": "ed25519",
   "headers": ["from","to","timestamp",
               "nonce","type","payload"],
   "signature": "..."}}
\end{verbatim}

\para{Signatures and keys.} The sending server canonicalizes the envelope with JCS~\cite{rfc8785} and signs it with Ed25519~\cite{rfc8032}.
The signature covers every field except \texttt{signature}, including unknown fields, and \texttt{headers} must list exactly those fields.
If an agent submits a message that already carries a signature, the server discards it.
\texttt{key\_id} names the DNS location of the key, \texttt{<selector>.atk.\_atp.<domain>}, where the key is published as a TXT record:
\begin{verbatim}
default.atk._atp.air.example. IN TXT
    "v=atp1 k=ed25519 p=<public key>"
\end{verbatim}

\para{Sender address policy (ATS).} A domain may publish a TXT record at \texttt{ats.\_atp.<domain>} listing the network addresses allowed to send for it, for example \texttt{v=atp1 deny=all allow=ip:\allowbreak{}192.0.2.0/24}.
Directives are applied in order, and later matches override earlier ones.
Receivers check only the TCP peer address and ignore self-reported forwarding headers.
Like SPF~\cite{rfc7208}, this is a cheap filter before signature verification.
In cloud deployments sending addresses may be shared, so it only plays a supporting role.
Without such a record the result is neutral, and the signature must always verify.

\para{Replay protection.} Receivers accept only timestamps within [now $-$ 300\,s, now $+$ 60\,s] and remember each (sender, \texttt{nonce}) pair seen in that window, rejecting duplicates.
If the replay store is full, the receiver returns a temporary error rather than accepting an unchecked message.

\para{Responses.} Table~\ref{tab:responses} lists the receiving domain's responses and how the sending domain handles them.

\begin{table}[t]
  \centering
  \caption{Responses of the receiving domain and handling by the sending domain.}
  \label{tab:responses}
  \ATPTableFont
  \begin{tabularx}{\columnwidth}{@{}>{\raggedright\arraybackslash}p{\ATPResponseLabelWidth}>{\raggedright\arraybackslash}X>{\raggedright\arraybackslash}p{\ATPResponseActionWidth}@{}}
    \toprule
    Response & Meaning & Sender action \\
    \midrule
    202 & Written to the recipient's inbox & Delivered \\
    Network error, 408, 429, 5xx & Temporary failure & Retry with backoff within the window \\
    400 \texttt{TIMESTAMP\_\allowbreak{}OUT\_\allowbreak{}OF\_\allowbreak{}WINDOW}, 504 \texttt{DEADLINE\_\allowbreak{}EXCEEDED} & Expired & Expired \\
    400 \texttt{NONCE\_\allowbreak{}REPLAYED} & Already accepted & Delivered if the last attempt was uncertain, else permanent failure \\
    Other 4xx (e.g., 403 bad signature, 404 unknown recipient) & Permanent failure & Stop \\
    \bottomrule
  \end{tabularx}
\end{table}

\begin{table*}[t]
  \centering
  \caption{Adversarial validation.
  Every case was run three times; the receiver's inbox was checked after each rejected request.}
  \label{tab:attacks}
  \ATPWideTable
  \begin{tabularx}{\textwidth}{@{}>{\raggedright\arraybackslash}p{\ATPAttackCaseWidth}>{\raggedright\arraybackslash}p{\ATPAttackConstructionWidth}>{\raggedright\arraybackslash}p{\ATPAttackCheckWidth}>{\raggedright\arraybackslash}X>{\raggedright\arraybackslash}p{\ATPAttackInboxWidth}c@{}}
    \toprule
    Case & Construction & Rejected by & Response & Inbox & Passed \\
    \midrule
    Forged signature & Sign with a random key & ATK signature verification & 403 \texttt{ATK\_\allowbreak{}SIGNATURE\_\allowbreak{}INVALID} & Absent & 3/3 \\
    Tampered content & Modify a captured payload & ATK signature verification & 403 \texttt{ATK\_\allowbreak{}SIGNATURE\_\allowbreak{}INVALID} & No new copy & 3/3 \\
    Sender/key domain mismatch & Sign noats.example as evil.example & Domain alignment & 403 \texttt{ATK\_\allowbreak{}SIGNATURE\_\allowbreak{}INVALID} & Absent & 3/3 \\
    Disallowed source address & Replay from attacker host & ATS policy & 403 \texttt{ATS\_\allowbreak{}VALIDATION\_\allowbreak{}FAILED} & No new copy & 3/3 \\
    Spoofed source headers & Forge X-Forwarded-For and X-Real-IP & ATS policy & 403 \texttt{ATS\_\allowbreak{}VALIDATION\_\allowbreak{}FAILED} & No new copy & 3/3 \\
    Replay & Resend the same sender/nonce & Replay cache & 400 \texttt{NONCE\_\allowbreak{}REPLAYED} & No new copy & 3/3 \\
    Expired timestamp & Timestamp 600 s in the past & Freshness window & 400 \texttt{TIMESTAMP\_\allowbreak{}OUT\_\allowbreak{}OF\_\allowbreak{}WINDOW} & Absent & 3/3 \\
    Future timestamp & Timestamp 600 s in the future & Freshness window & 400 \texttt{TIMESTAMP\_\allowbreak{}OUT\_\allowbreak{}OF\_\allowbreak{}WINDOW} & Absent & 3/3 \\
    \bottomrule
  \end{tabularx}
\end{table*}

\para{Retry and expiry.} The retry interval starts at 1\,s and doubles.
When a message's age approaches the 300\,s window (with a default margin of 10\,s), or the next retry would fall outside it, the sending domain stops and marks the message as expired.
Expired or permanently failed messages are bounced to the sending agent only if the \texttt{payload} sets \texttt{ack\_required}.
Otherwise they are dropped and counted.
Inboxes keep messages until a domain-configured retention period ends.
Figure~\ref{fig:message-delivery} summarizes the delivery sequence and the handling of unavailability and expiry.

\para{Request/response.} A response's \texttt{in\_reply\_to} names the request's \texttt{nonce}.
A request's deadline is its \texttt{timestamp} plus the \texttt{timeout} in the \texttt{payload}, or the end of the freshness window if no timeout is given.
Every server computes the deadline the same way: a request that arrives past its deadline gets 504, and a queued request past its deadline is marked expired.
The requesting agent waits with a long poll, and the server returns only messages whose \texttt{in\_reply\_to} matches and that come from the original recipient.
Figure~\ref{fig:request-response} shows the request/response exchange, deadline checks, and response matching.

\para{Event subscription.} A subscription is itself a request whose \texttt{payload} carries a \texttt{subscribe} action, a subscription identifier, optional event types, and a lifetime (default 3600\,s).
The same subscriber renews or cancels with the same identifier.
The publisher's server checks that the subscription is valid at submission and before each delivery and does not deliver events of expired subscriptions.
It sends heartbeats at a fixed interval (default 60\,s), which the subscriber's server records without placing them in the agent's inbox.
Figure~\ref{fig:publish-subscribe} shows the subscription lifecycle and event delivery.

\section{Security Tests}
\label{app:security}

We built malicious messages for the assumptions and threat model in Section~\ref{sec:atp_design} and sent them from one server to another (Table~\ref{tab:attacks}).
Each message violates only the property under test and is otherwise valid, so that a rejection comes from the intended check.
For each case we checked the status code, the error code, and that the recipient's inbox did not gain the message.
When a case resends a captured legitimate message, we checked that the inbox still holds only the original, unchanged copy.
The sending domain \texttt{noats.example} publishes a signing key but no ATS record, since ATS is optional, so the forged-signature and domain-mismatch cases can only be stopped by signature checks.

All eight cases were rejected by the intended check in each of three runs.
The domain-mismatch case deserves attention: the attacker signs with a real, valid key that belongs to another domain, and without the alignment check the message would be accepted.
These tests show that the implementation rejects these inputs as designed.
They do not prove the protocol secure in general.

\section{Delivery Bound When the Receiving Domain Is Unreachable}
\label{app:outage}

Section~\ref{sec:atp_design} notes that the sending domain judges freshness by the message's creation time, so server-to-server store-and-forward can only tolerate a bounded outage of the peer.
This appendix computes the bound from the retry rule and checks the implementation against it.

\para{Analytical bound.} The retry interval starts at 1\,s and doubles.
The sending domain stops when a message's age approaches the 300\,s window (minus a 10\,s margin, i.e., 290\,s) or when the next retry would fall outside it.
If connections are refused immediately during the outage, a message sent at the start of the outage is attempted at ages of about 0, 1, 3, 7, 15, 31, 63, 127, and 255\,s.
The next attempt would fall at about 511\,s, beyond the window, so the message expires after the attempt at about 255\,s fails.
The duration of each attempt shifts these points slightly later.
If the peer stops answering instead of refusing, each attempt waits for a timeout (30\,s by default) and the points move much later, so the calculation no longer applies.

\para{Setup.} We made the receiving server's \sys port reply to new connections with a TCP RST while keeping the server process and container running, so every delivery attempt fails at once, as the calculation assumes.
The sending domain sent one message every 5\,s.
The outage lasted 30\,s, 240\,s (15\,s before the bound), or 270\,s (15\,s after the bound), with three runs per duration.
At the end, we classified each message as delivered, expired, or pending, recorded its attempts, and compared it with the prediction from the retry rule.

\para{Results.} With a 30\,s outage, all 10 messages were delivered in every run.
With 240\,s, all 52 messages were delivered in every run.
With 270\,s, 54 to 55 messages were delivered and 3 to 4 expired per run, and all expired messages had been sent within 15\,s after the outage began.
No run had duplicates, permanent failures, or pending messages.
One message differed from the prediction: it was sent 15.03\,s into the outage, and the rule placed its last attempt 29\,ms before the outage ended, but the time spent in earlier attempts pushed the last attempt past the end of the outage, so it was delivered.
This matches the expected shift caused by attempt durations.
Figure~\ref{fig:outage} plots each message's outcome against the computed retry schedule.
After recovery, messages wait for their next scheduled retry rather than being sent immediately.
In the 240\,s group the median wait after recovery was 36\,s and the maximum was 125\,s.

\balance
This bound only limits forwarding between servers.
The window is checked only at cross-domain receipt, and once a message is in the receiving inbox, the server no longer checks its timestamp when an agent fetches it.
The case study and the mechanism checks in Section~\ref{sec:eval} fetched messages after about 360\,s and 600\,s offline.

\end{document}